\documentclass{article}

\usepackage{makecell}

\usepackage{graphics}
\usepackage[table]{xcolor}
\usepackage{graphicx}
\usepackage{subcaption}
\usepackage{caption}
\usepackage{amssymb}
\usepackage{algorithm}
\usepackage{algpseudocode}
\usepackage{wrapfig}
\usepackage{longtable}
\usepackage{fancyvrb}
\usepackage{empheq}
\definecolor{metacolor}{HTML}{0064E0}

\usepackage{adjustbox}
\usepackage{makecell} 
\definecolor{color_blue}{HTML}{E7EFFA}
\definecolor{color_green}{HTML}{E6F8E0}
\definecolor{color_gray}{HTML}{ECECEC}
\definecolor{pearDark}{HTML}{2980B9}

\usepackage[preprint]{neurips_2025}

\usepackage[utf8]{inputenc} 
\usepackage[T1]{fontenc}    
\usepackage{hyperref}       
\usepackage{url}            
\usepackage{booktabs}       
\usepackage{amsfonts}       
\usepackage{nicefrac}       
\usepackage{microtype}      
\usepackage{xcolor}         
\usepackage{tabularx}
\usepackage{multirow}
\usepackage{graphicx}
\usepackage{amsmath}

\title{UNIC: Unified In-Context Video Editing}

\author{%
  Zixuan Ye\textsuperscript{1}\thanks{Equal contribution. Work done during an internship at KwaiVGI, Kuaishou Technology. }\hspace{0.7em}
  Xuanhua He\textsuperscript{1}\footnotemark[1]\hspace{0.7em}
  Quande Liu\textsuperscript{2}\thanks{Corresponding author.}\hspace{0.7em}
  Qiulin Wang\textsuperscript{2}\hspace{0.7em}
  Xintao Wang\textsuperscript{2}\hspace{0.7em} \\
  \textbf{Pengfei Wan}\textsuperscript{\textbf{2}}\hspace{0.7em} 
  \textbf{Di Zhang}\textsuperscript{\textbf{2}}\hspace{0.7em}
  \textbf{Kun Gai}\textsuperscript{\textbf{2}}\hspace{0.7em}
  \textbf{Qifeng Chen}\textsuperscript{\textbf{1}}\hspace{0.7em}
  \textbf{Wenhan Luo}\textsuperscript{\textbf{1}}\footnotemark[2] \\ 
  \textsuperscript{1}The Hong Kong University of Science and Technology \\
  \textsuperscript{2}Kuaishou Technology\\
  \texttt{\href{https://zixuan-ye.github.io/UNIC}{https://zixuan-ye.github.io/UNIC}}
}

\begin{document}

\maketitle

\begin{abstract}

Recent advances in text-to-video generation have sparked interest in generative video editing tasks. Previous methods often rely on task-specific architectures (e.g., additional adapter modules) or dedicated customizations (e.g., DDIM inversion), which limit the integration of versatile editing conditions and the unification of various editing tasks. In this paper, we introduce UNified In-Context Video Editing (UNIC), a simple yet effective framework that unifies diverse video editing tasks within a single model in an in-context manner. To achieve this unification, we represent the inputs of various video editing tasks as three types of tokens: the source video tokens, the noisy video latent, and the multi-modal conditioning tokens that vary according to the specific editing task. Based on this formulation, our key insight is to integrate these three types into a single consecutive token sequence and jointly model them using the native attention operations of DiT, thereby eliminating the need for task-specific adapter designs. Nevertheless, direct task unification under this framework is challenging, leading to severe token collisions and task confusion due to the varying video lengths and diverse condition modalities across tasks. To address these, we introduce task-aware RoPE to facilitate consistent temporal positional encoding, and condition bias that enables the model to clearly differentiate different editing tasks. This allows our approach to adaptively perform different video editing tasks by referring the source video and varying condition tokens "in context", and support flexible task composition. To validate our method, we construct a unified video editing benchmark containing six representative video editing tasks. Results demonstrate that our unified approach achieves superior performance on each task and exhibits emergent task composition abilities. 

\end{abstract}

\begin{figure}
    \centering
    \includegraphics[width=1\linewidth]{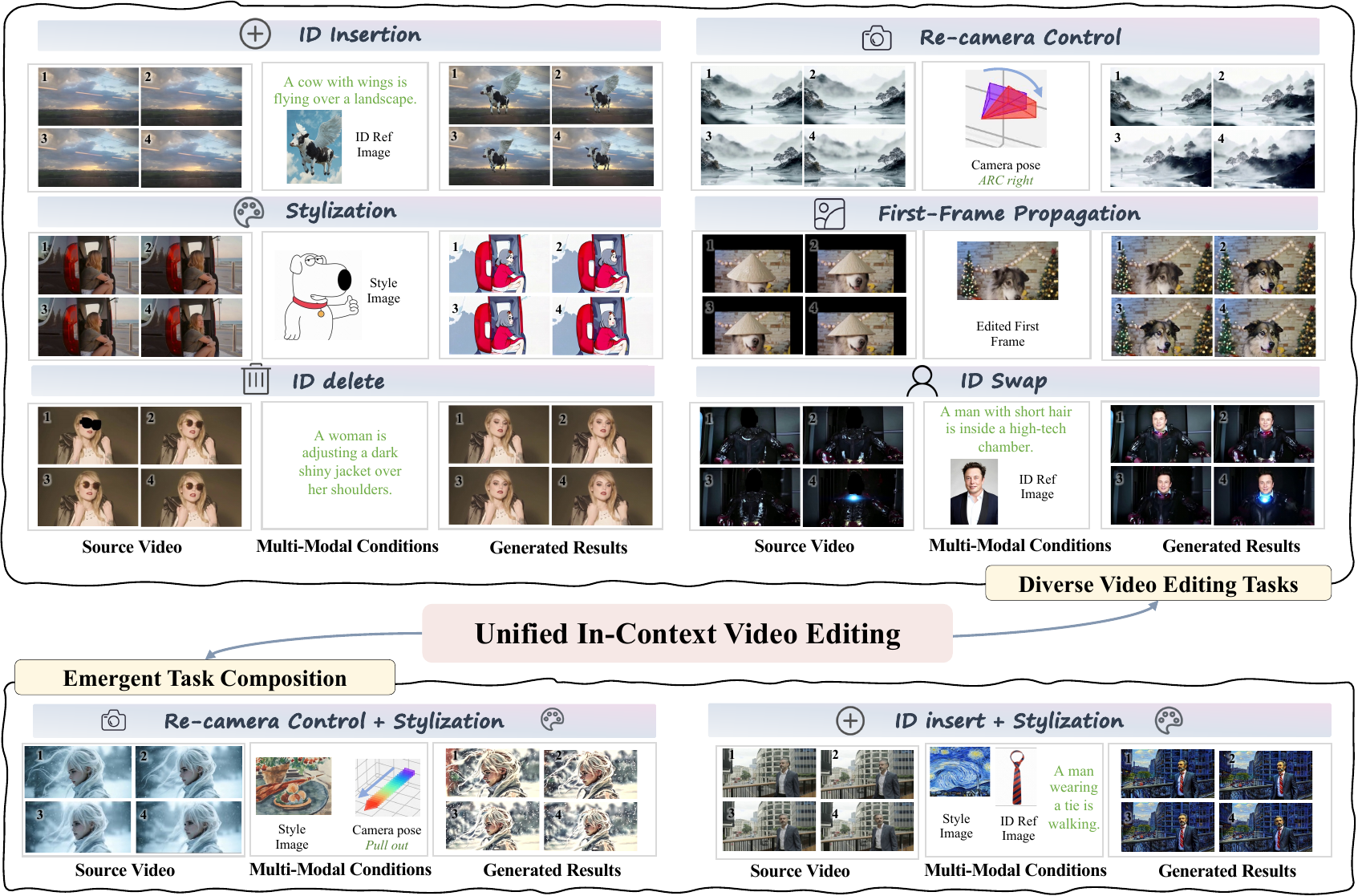}
    \caption{\textbf{Unified In-Context Video Editing enables unified video editing and emergent task composition.} Here we demonstrate the unification of six representative tasks, including ID Insert/Delete/Swap, Re-Camera Control, Stylization, and Propagation. }
    \label{fig:teaser}
\end{figure}
\vspace{-10pt}

\section{Introduction}

Recent years have witnessed significant advancements in text-to-video foundation models based on diffusion~\cite{videocrafter1, wan2025, cogvideo, yang2024cogvideox,hacohen2024ltx,klingai2025}, establishing powerful tools for creating video content. Beyond the generation, video editing emerges as a natural extension, which aims to re-generate a reference video under multi-modal conditions, often incorporating fine-grained control signals like subject ID~\cite{he2024id,zhang2025magic,yuan2024identity} and artistic style~\cite{stylecrafter} along with textual prompts. Video editing spans diverse tasks, including global editing like style transfer~\cite{stylecrafter,ye2024stylemaster}, video propagation~\cite{liu2024generative,anyv2v}, and local editing like object insertion, removal, swap~\cite{tu2025videoanydoor,zi2025cococo,lee2025video}, as well as video re-rendering like re-camera control~\cite{bai2025recammaster}. These tasks hold vast potential for applications in film production, virtual reality, and automated content creation.

\begin{figure}[!t]
    \centering
    \includegraphics[width=1\linewidth]{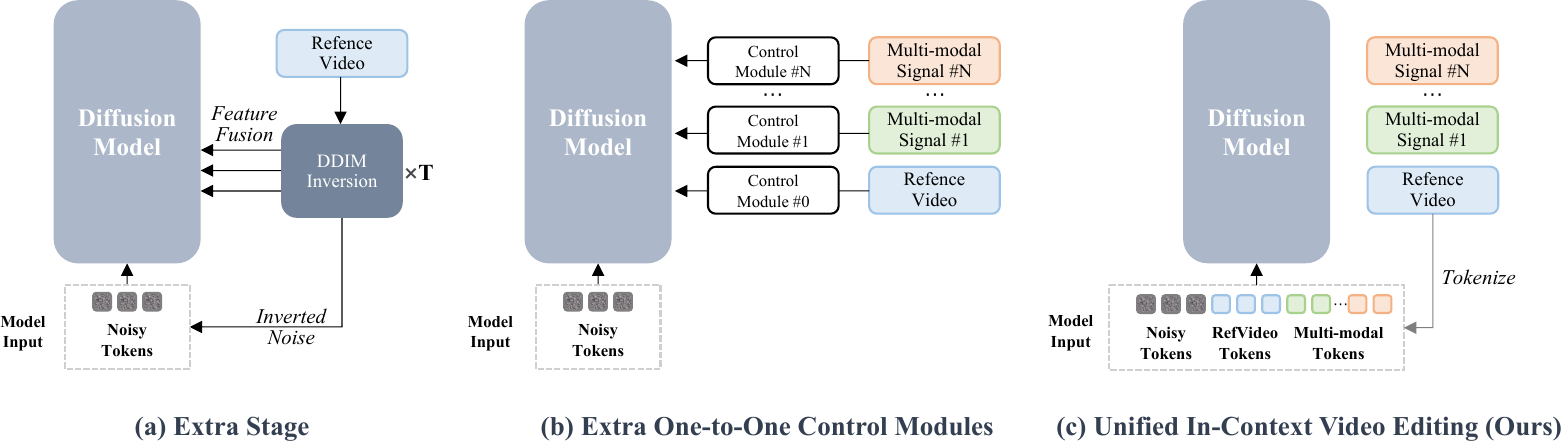}
    \caption{\textbf{Architectural comparison for incorporating conditioning signals.} \textbf{(a) Extra Stage:} Utilizes DDIM inversion on a reference video to derive inverted noise. \textbf{(b) Extra One-to-One Control Modules:} Employs dedicated, separate modules to process each control signal (e.g., reference video, multi-modal signals) and inject guidance into the diffusion model. \textbf{(c) In-Context Video Editing (Ours):} Our proposed method directly integrates guidance by tokenizing all conditioning signals (reference video, multi-modal signals) and concatenating them with the noisy input tokens, allowing the diffusion model to process all information jointly within its input sequence.}
    \label{fig:arc_comparison}
\end{figure}

Current video editing methods primarily follow two strategies to inject reference video and control signals. As depicted in Fig.~\ref{fig:arc_comparison}, one stream of methods, represented by Video-P2P~\cite{liu2024video}, AnyV2V~\cite{anyv2v}, and FLATTEN~\cite{flatten}, utilizes DDIM inversion for noise initialization to preserve the main structure of the reference video. However, these methods often fail to achieve ideal results and will inevitably introduce an additional stage, doubling the inference steps and cost. Another stream generally employs adapter-based designs~\cite{anyv2v,zi2025cococo,bai2025recammaster,jiang2025vace,wang2024videocomposer,zhang2023controlvideo,mou2024revideo} to inject different conditions, including reference video and multiple control signals. Despite promising progress, these methods suffer from two main challenges: 1) the adapter-based designs require modification to the model architectures and introduce parameter redundancy; and 2) these methods are generally task-specific, requiring training separate modules for each condition signal, raising difficulty for task extendability and unification. Very recently, VACE~\cite{jiang2025vace} tries to categorize condition signals into frames and masks for unified video editing, yet still requires heavy adapter designs and is limited to process only visual conditions.

Based on these problems, this paper presents a unified and efficient framework for video editing tasks from multi-modal signals, named UNified In-Context Video Editing (UNIC). 
Inspired by the recent advancements in large language models and visual content generation~\cite{yang2024qwen2,bai2024longbench,bai2025qwen2,wang2025internvideo2,chen2024longvila,song2025insert,tan2024ominicontrol,xiao2024omnigen}, our key insight is to integrate diverse input signals from various editing tasks as a combined token sequence along the frame dimension, which are jointly modeled using the native transformer attentions to learn editing tasks from diverse context conditions. As shown in Fig.~\ref{fig:arc_comparison}, to achieve unified video editing, UNIC formulates the inputs of different video editing tasks as three kinds of tokens, i.e.,  \textit{1) the VAE tokens from reference video}, \textit{2) the multi-modal condition signals} that vary upon the editing tasks, as well as \textit{3) the noisy video latent}. By jointly concatenating these tokens and dynamically varying the condition token ``in context", UNIC can flexibly perform diverse editing tasks without any task-specific architectural changes. 



Crucially, directly concatenating these diverse input tokens presents unavoidable challenges for unified video editing. Firstly, the multi-modal conditions from different task types present inconsistent lengths, raising difficulty in achieving correct alignment with the video. For example, camera poses usually have a direct frame-to-frame correspondence to each video frame, while the style images directly affect the entire video. 
Such inconsistency makes it challenging to deal with varying-length video editing and leads to inevitable index collisions. 
Therefore, we propose Task-aware RoPE, which dynamically assigns unique Rotary Positional Embedding (RoPE) indices based on different task types, ensuring coherent temporal understanding regardless of varying condition length. Furthermore, different editing tasks may share the same modality of conditions (e.g., an image may represent an object identity in object editing or style in video stylization), leading to task confusion. To this end, we introduce a learnable condition bias for multi-modal condition signals, which enables the model to adaptively learn the target task type and resolve task ambiguity. 

To validate the performance of our proposed unified framework, we construct a unified video editing benchmark incorporating six representative video editing tasks with distinct editing area ratio and conditioning modalities, including: local editing tasks of ID Swap/Delete/Insert~\cite{jiang2025vace,bian2025videopainter}; global editing tasks of stylization~\cite{ye2024stylemaster} and propagation~\cite{liu2024generative}; and the re-rendering task of re-camera control~\cite{bai2025recammaster}. These tasks exhibit significant variation in their input modalities (including text, images, and camera poses). Experimental results indicate that, despite the wide range of differences among these tasks, our framework not only successfully unifies them but also delivers superior performance across all. As shown in Fig.~\ref{fig:teaser}, our method offers two distinct advantages: 1) supporting a variety of editing tasks within a single framework without necessitating architectural changes, showcasing high flexibility; and 2) emergent capability to combine various editing tasks, highlighting its potential to unlock more complex and creative editing possibilities. We also provide in-depth analysis for unified video generation, respectively about the advantages of unified training over single task training as well as the training order of different tasks. 

\section{Related Work}

\subsection{Video Editing and Re-rendering}
Video editing encompasses diverse tasks, ranging from local adjustments to global transformations of a reference video. Achieving these modifications often requires injecting multi-modal signals, such as motion~\cite{tu2025videoanydoor,wang2024motionctrl,wang2025cinemaster,xiao20243dtrajmaster,wei2024dreamvideo}, style~\cite{stylecrafter,ye2024stylemaster}, object attributes~\cite{tu2025videoanydoor,zi2025cococo,huang2025conceptmaster,he2024id,wei2024dreamvideo,zhang2025magic}, or camera pose~\cite{bai2025recammaster,bai2024syncammaster,wang2024motionctrl}, into the generation process.

To preserve information from the reference video during editing, several methods employ DDIM inversion. This technique initializes the generation noise based on the reference video and injects features extracted during the inversion process into the denoising steps. For instance, VideoP2P~\cite{liu2024video} copies inversion features and replaces specific cross-attention maps to align with editing requirements. Similarly, FLATTEN~\cite{flatten} utilizes optical flow to identify keypoints and injects their features to maintain motion fidelity. AnyV2V~\cite{anyv2v} also leverages spatial, temporal, and CNN features gathered during inversion. While these approaches excel at retaining reference video information, they inherently require an \textbf{additional stage} for inversion, thereby increasing the overall inference cost and computational overhead.

Instead of additional processing stages, another strategy focuses on injecting control directly into the denoising network using auxiliary modules. These modules aim to preserve or guide specific aspects based on reference video or multi-modal signals. To maintain structural or layout information of reference videos, Follow-your-Canvas~\cite{chen2024follow} extracts window details with a layout encoder, while MagicEdit~\cite{liew2023magicedit} leverages the depth video of the reference video via a depth ControlNet~\cite{controlnet}.  For finer-grained preservation of content and motion details, VideoAnyDoor~\cite{tu2025videoanydoor} and Revideo~\cite{mou2024revideo} use separate encoders to obtain the feature and inject them through ControlNet. To incorporate additional multi-modal conditions, VideoAnyDoor employs an extra ID encoder, infusing identity information via cross-attention. Likewise, StyleMaster~\cite{ye2024stylemaster} introduces a dedicated style encoder, injecting style features through cross-attention. A common choice of these methods is the reliance on specialized control modules for each condition type. This design choice, requiring \textbf{additional control modules for different conditions}, inevitably increases overall model complexity and limits the extensibility to diverse or novel video editing tasks.

\subsection{Universal Generative Models}
Developing unified ``omni'' solutions capable of handling diverse generative tasks within a single model is a challenging but highly valuable goal. The field of image editing and generation has witnessed a clear trend towards such unification. Early works like Instruct-imagen~\cite{hu2024instruct} integrated multi-modal instructions using cross-attention. OmniGen~\cite{xiao2024omnigen} further advances this by tokenizing all conditions as direct inputs to the transformer, creating a flexible "any-purpose" generation model without external plugins. This powerful approach of handling conditions directly within the core architecture has been validated and extended by subsequent research, including OminiControl~\cite{tan2024ominicontrol}, ACE~\cite{han2024ace}, UniReal~\cite{chen2024unireal}, and other unified models~\cite{le2024one,song2025insert,mou2025dreamo}.

In contrast, unified approaches in video generation and editing often still depend on task-specific control mechanisms. VideoComposer~\cite{wang2024videocomposer}, for instance, employs different ControlNets for various inputs, and VACE~\cite{jiang2025vace} uses specialized control blocks to inject conditional information. Drawing inspiration from recent image editing paradigms that successfully leverage full-attention mechanisms to replace dedicated modules like ControlNet and adapters, we propose to extend this strategy to video editing. Although FullDiT~\cite{ju2025fulldit} has demonstrated the potential of 3D full-attention for multi-control video generation, the leap from generation to more difficult video editing represents a distinct and open challenge. Our work aims to fill this gap by \textbf{developing a unified and
flexible method for general video editing purposes}, eliminating the need for separate control modules.

\section{Method}
Towards general video editing, we first review the diverse video editing tasks, and systematically define all inputs across different tasks into three basic types. Building on this, we introduce an in-context video editing framework, offering a parameter-efficient and highly flexible approach adaptable to various editing purposes. We further elaborate on the specific design within our architecture for task differentiation and flexibility.


\subsection{Multi-modal Driven Video Editing Tasks}
\label{sec:token_types}
\begin{figure}
    \centering
    \includegraphics[width=1\linewidth]{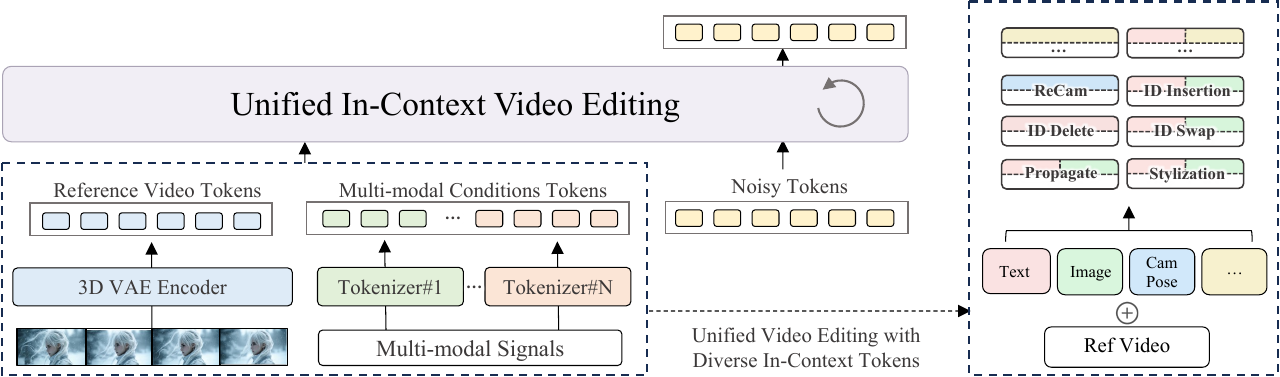}
    \caption{\textbf{Overall Pipeline of Unified In-Context Video Editing.} Our framework utilizes a unified transformer architecture for video editing. The model input is created by concatenating noisy tokens, reference video tokens, and multi-modal condition tokens (task-specific controls like images), these combined tokens form a single input sequence along the frame dimension. By simply modifying the multi-modal condition tokens, this framework can handle any video editing task.}
    \label{fig:pipeline}
\end{figure}

Diverse video editing tasks, as illustrated on the right of Fig.~\ref{fig:pipeline}, incorporate different condition inputs. For example, stylization requires reference video and style image; object insertion needs reference video, text and object image; re-camera control requires reference video and camera poses. In summary, these tasks are fundamentally driven by unique combinations of multi-modal inputs. Therefore, we generalize these inputs into three basic types: \textbf{Noisy tokens}, \textbf{Reference video tokens}, and \textbf{Multi-modal condition tokens}. This classification enables any video editing task to be represented and processed within our structured approach.

\textbf{Noisy tokens} represent the initial latent state of the target video, typically derived from random noise or a noise-added input video latent. 

\textbf{Reference video tokens} represent the VAE tokens of reference video, providing essential temporal context, motion dynamics, and visual content within the reference video. The influence of these tokens varies with specific task's alignment requirements: 1) \textbf{Strict-Alignment:} For tasks like ID delete or stylization, these tokens enforce strong frame-by-frame correspondence, ensuring the output precisely follows the motion and unedited content; 2) \textbf{Soft-Reference:} For tasks like re-camera control, these tokens provide more abstract guidance, following the overall content or motion style but without demanding strict pixel-level matching, allowing for significant deviations.

\textbf{Multi-modal condition tokens} include all other forms of guidance signals. This versatile category includes, but is not limited to:
\begin{itemize}
    \item \textbf{Image Tokens:} encoding reference images, guiding either structural edits like the edited first frame in the video propagation task, or serving as style reference image in stylization and ID reference image in ID insertion.
    \item \textbf{Auxiliary Control Tokens:} Representing diverse signals like camera pose, depth maps, segmentation masks, human pose skeletons, edge maps, sparse trajectories for motion guidance, audio signals for lip-syncing, or other structural/geometric constraints for fine-grained control.
\end{itemize}

By classifying all inputs into these three token types, the specific combination of reference video tokens and multi-modal condition tokens can represent any video editing task, making it easier to uniformly process these tasks, serving as the foundation for a unified framework.

\subsection{Unified In-Context Video Editing}
We introduce Unified In-Context Video Editing (UNIC), a simple yet effective paradigm for diverse video editing tasks. The core idea is to represent all inputs, including the noisy tokens, the reference video, and all other multi-modal conditioning signals (images, cameras, etc.), as a single, unified token sequence. This contrasts with approaches that inject conditions via additional control modules. By processing all information jointly within the full-attention layers, the model can flexibly learn the \textbf{contextual relationships} for different editing tasks from the conditions provided ``in context''.

\paragraph{Preliminary.}

Our method inherits the video diffusion transformers trained using flow matching. Specifically, the training objective is given by:
\begin{equation}
\label{eq:flow_matching}
\mathcal{L}_\textrm{FM}(\theta) = \mathbb{E}_{t, \boldsymbol{x}_0, \boldsymbol{x}_1} \left\|\boldsymbol{v}_\theta(\boldsymbol{x}_t,t) - (\boldsymbol{x}_1-\boldsymbol{x}_0) \right\|_2^2,
\end{equation}
where $\boldsymbol{x}_1\sim p(\boldsymbol{x}_1)$ represents the video sample, $\boldsymbol{x}_0\sim\mathcal{N}(\boldsymbol{0},\boldsymbol{1})$ is a Gaussian sample, and $t$ is randomly distributed in $[0,1]$. The function $\boldsymbol{v}_\theta$ is the neural network that takes the noised version $\boldsymbol{x}_t=t \boldsymbol{x}_1 + (1 - t) \boldsymbol{x}_0$ as input.

Training with this loss function leads to the following ordinary differential equation (ODE):
\begin{equation}
\frac{d\boldsymbol{x}_t}{dt}=\boldsymbol{v}_\theta(\boldsymbol{x}_t,t),
\end{equation}
which allows us to sample a synthesized video $\boldsymbol{x}_1$ from a random Gaussian noise $\boldsymbol{x}_0$.

Our transformer consists of several DiT blocks, where each block contains 2D self-attention to learn spatial information and 3D self-attention to fuse spatio-temporal information.

\subsubsection{In-Context Video Editing}
For video editing and re-rendering tasks, given a reference video $V_{ref} \in \mathbb{R}^{f \times c \times h \times w}$ and a set of conditions $\{C_i\mid i=1 \dots n\}$, our goal is to generate a target video $V_{tar} \in \mathbb{R}^{f \times c \times h \times w}$ that aligns with these conditions while preserving required content of $V_{ref}$.

Existing methods often preserve reference video information and inject conditions either through DDIM inversion for noise initialization~\cite{anyv2v,flatten} or add additional control modules like ControlNet~\cite{ye2024stylemaster,zhang2023controlvideo}.  Instead, we propose a unified and parameter-efficient approach, concatenating all the inputs along the frame dimension to perform self-attention. First, we encode the reference video $V_{ref}$ using a 3D VAE encoder to obtain its latent representation $z_{ref}$. Similarly, other conditions $C_i$ are converted into token sequences $z_i$ using modality-specific tokenizers (e.g., the same 3D VAE for image conditions, a T5 tokenizer for text, an MLP for camera pose, etc.).

As shown in Fig.~\ref{fig:pipeline}, during generation, the model operates on a noisy latent $z_{tar}$ (which starts as random noise and evolves towards the final latent representation of the target video). By concatenating multi-modal tokens as: $z_{cond} = [z_1; \ldots; z_N]$, which is then combined with noisy token $z_{tar}$, reference video tokens $z_{ref}$ in frame dimension into a single sequence as the model input $z = [z_{tar}; z_{ref}; z_{cond}]$, we can simply perform full 3D attention to enable the interaction of the tokens. Without requiring an intricate inversion process or task-specific architectural modifications, the only requirements for this framework are the modality-specific tokenizers, which are also needed in other methods.


\subsubsection{Effective and Flexible Task Unification}
Based on in-context editing framework, theoretically we can unify all the video editing tasks. However, directly concatenating them introduces specific challenges: \textbf{1) Task Ambiguity:} When different editing tasks rely on conditions from the same modality (an image can refer to style reference or ID), simple concatenation can make it difficult for the model to distinguish the target task of the conditional tokens. \textbf{2) Positional Encoding Conflicts \& Inflexibility:} The 3D RoPE used in video generation base models, often assigns sequential indices in the frame dimension, which will raise problems when handling different conditions. For example, style reference does not have a direct correspondence with video frames, whereas camera pose requires frame-to-frame correspondence, making it difficult to maintain appropriate alignments under sequential indices. The situation becomes more problematic in variable-length editing, where it even struggles to clearly distinguish boundaries between reference videos and conditions. To address these difficulties, we introduce two key components: Condition Bias and Task-Aware RoPE Index.

\paragraph{Condition Bias}
As mentioned above, to address the task ambiguity, we propose \textit{Condition Bias}, the task-specific learnable embedding that is directly added to tokens before attention computation. To be specific, for each condition in the context, including the multi-modal signals and reference video, i.e., $z_i \in \{z_{\text{ref}}, z_1, \ldots, z_N\}$, we inject a learnable bias $b_i \in \mathbb{R}^d$ corresponding to its task type:
\begin{equation}
    \tilde{z}_i = z_i + b_i\,.
\end{equation}
These task-aware tokens $\{\tilde{z}_i\}$ then undergo standard full self-attention. The biases implicitly guide attention by structuring token representations: tokens from the same task share similar bias-induced feature offsets, promoting intra-task alignment while maintaining cross-task distinction. The learnable embeddings are zero-initialized to preserve original token semantics, and their dimension-preserving addition enables simple integration with existing architectures.

\paragraph{Task-aware RoPE Index}
Standard 3D Rotary Position Embedding (RoPE) assigns sequential indices to frames. For instance, in a re-camera control task with $N$ frames, noisy tokens might occupy indices $0$ to $N-1$, reference video tokens $N$ to $2N-1$, and camera poses $2N$ to $3N-1$. When the video frame length $N$ varies, it will be hard to find the boundaries between the conditions, causing poor frame alignment.

To address this, we introduce a task-aware RoPE indexing scheme: 1) For tasks where conditional inputs have a direct frame-to-frame correspondence with the video (e.g., reference videos, camera poses, audio), we reuse the indices of the noisy latent video frames ($0$ to $N-1$), which can help maintain the alignment. 2) For tasks that do not have such direct frame-to-frame correspondence (e.g., ID images, style references), we assign indices based on a base offset and task-specific offset. The base offset, $m$, is determined by the video length (i.e., $m = N$). Beyond this, each task $t$ is assigned with a pre-defined, fixed task offset, $O_t$, and a slot count (or length), $L_t$. The value $O_t$ dictates the starting point of task $t$'s allocation relative to $m$, and $L_t$ defines how many consecutive indices the task occupies.
The index range for such a task $t$ is calculated as:
\begin{equation}
\text{Index}(t) = (m + O_t) + [0, \dots, L_t-1]
\end{equation}
These task offsets $O_t$ and lengths $L_t$ are chosen to ensure non-overlapping task slots and maintain clear task distinction. For example, if ID images (task 1) are defined with a task offset $O_1=100$ and require $L_1=3$ slots (supporting up to $3$ IDs), they will use indices from $N+100$ to $N+102$. If a style reference (task 2) is defined with a task offset $O_2=200$ and requires $L_2=1$ slot, it will occupy the index $N+200$. This adaptive approach ensures that task slots automatically scale with the video length $N$, while preserving their relative positional relationships and avoiding overlap.

\begin{table}[!t]
\tiny
\centering
\renewcommand{\arraystretch}{1.15}
\caption{Quantitative comparison on six video editing tasks: ID Insert/Swap/Delete, Re-Camera control, Stylization, and Propagation. Best results are highlighted in \textbf{bold}. $\uparrow$ indicates higher is better;  $\downarrow$ indicates lower is better.
}
\label{maintable}
\resizebox{\textwidth}{!}{
\begin{tabular}{@{}l|cc|cccc@{}}
\hline
\multicolumn{7}{c}{\textbf{ID Insert}}                                                                                                                                                                    \\ \hline
{\multirow{2}{*}{\textbf{Method}}} & \multicolumn{2}{c|}{\textbf{Identity}}                     & {\textbf{Alignment}}  & \multicolumn{3}{c}{\textbf{Video Quality}}                  \\
& \textbf{CLIP-I$\uparrow$} & {\textbf{DINO-I$\uparrow$}} & {\textbf{CLIP-score$\uparrow$}} & \textbf{Smoothness$\uparrow$} & \textbf{Dynamic$\uparrow$} & \textbf{Aesthetic$\uparrow$} \\ \hline
VACE~\cite{jiang2025vace}& 0.522& 0.110& 0.100& 0.933& \textbf{44.568}& 5.407\\
Ours& \textbf{0.598}& \textbf{0.245}& \textbf{0.216}& \textbf{0.961}& 11.07& \textbf{5.627}\\ \hline
\multicolumn{7}{c}{\textbf{ID Swap}} \\ \hline
VACE~\cite{jiang2025vace}& 0.712& 0.423& 0.230& 0.964&\textbf{29.306}& 6.015\\
AnyV2V(Prop)~\cite{anyv2v}& 0.605& 0.229& 0.218& 0.917& 7.596& 4.842\\
Ours(Prop)& 0.693& 0.414& 0.236& \textbf{0.980}& 5.153& 5.801\\
Ours& \textbf{0.725}& \textbf{0.429}& \textbf{0.242}& 0.971& 7.500& \textbf{6.056}\\ \hline
\multicolumn{7}{c}{\textbf{ID Delete}}                                                                                                                                                                    \\ \hline
{\multirow{2}{*}{\textbf{Method}}} & \multicolumn{2}{c|}{\textbf{Video Reconstruction}}         & {\textbf{Alignment}}  & \multicolumn{3}{c}{\textbf{Video Quality}}                  \\
& \textbf{PSNR}$\uparrow$ & {\textbf{RefVideo-CLIP$\uparrow$}}& {\textbf{CLIP-score$\uparrow$}} & \textbf{Smoothness$\uparrow$} & \textbf{Dynamic$\uparrow$} & \textbf{Aesthetic$\uparrow$} \\ \hline
AnyV2V(Prop)~\cite{anyv2v}& 19.504& 0.869& 0.205& 0.964& 4.980& 5.325\\
VACE~\cite{jiang2025vace}& 20.947& 0.883& 0.211& 0.966& \textbf{15.441}& 5.332\\
VideoPainter~\cite{bian2025videopainter}& \textbf{22.987}& \textbf{0.920}& 0.212& 0.957& 13.759& 5.403\\
Ours(Prop)& 20.378& 0.906& 0.209& 0.968& 9.017& 5.408\\
Ours& 19.171& 0.900& \textbf{0.217}& \textbf{0.970}& 10.934& \textbf{5.493}\\ \hline

\multicolumn{7}{c}{\textbf{Propagation}}                                                                                                                                                          \\ \hline
{\multirow{2}{*}{\textbf{Method}}} & \multicolumn{2}{c|}{\textbf{Frame Alignment}}               & {\textbf{Alignment}}  & \multicolumn{3}{c}{\textbf{Video Quality}}                  \\
& \multicolumn{2}{c|}{{\textbf{RefVideo-CLIP$\uparrow$}}}& {\textbf{CLIP-score$\uparrow$}} & \textbf{Smoothness$\uparrow$} & \textbf{Dynamic$\uparrow$} & \textbf{Aesthetic$\uparrow$} \\ \hline
AnyV2V~\cite{anyv2v}& \multicolumn{2}{c|}{0.812}& 0.229& 0.935& 13.462& 5.136\\
VACE(I2V)~\cite{jiang2025vace}& \multicolumn{2}{c|}{0.574}& \textbf{0.234}& 0.932& \textbf{36.783}& 5.425\\
Ours& \multicolumn{2}{c|}{\textbf{0.840}}& 0.226& \textbf{0.966}& 12.762& \textbf{5.565}\\ \hline

\multicolumn{7}{c}{\textbf{Stylization}}                                                                                                                                                               \\ \hline
{\multirow{2}{*}{\textbf{Method}}} & \multicolumn{2}{c|}{\textbf{Style \& Content}}               & {\textbf{Alignment}}  & \multicolumn{3}{c}{\textbf{Video Quality}}                  \\
& \textbf{CSD-Score$\uparrow$} & {\textbf{ArtFID$\downarrow$}} & {\textbf{CLIP-score$\uparrow$}} & \textbf{Smoothness$\uparrow$} & \textbf{Dynamic$\uparrow$} & \textbf{Aesthetic$\uparrow$} \\ \hline
AnyV2V(Prop)~\cite{anyv2v}& 0.207& 43.299& 0.195& 0.937& 9.227& 4.640\\
StyleMaster~\cite{ye2024stylemaster}& \textbf{0.306}& 38.213& 0.188& \textbf{0.952}& 9.758& 5.121\\
Ours(Prop)& 0.197& \textbf{36.198}& \textbf{0.215}& 0.932& \textbf{11.569}& 5.045\\
Ours& 0.259& 37.619& 0.171& 0.945& 9.370& \textbf{5.276}\\\hline
\multicolumn{7}{c}{\textbf{Re-Camera Control}}                                                                                                                                                          \\ \hline
{\multirow{2}{*}{\textbf{Method}}} & \multicolumn{2}{c|}{\textbf{Camera Control}}               & {\textbf{Alignment}}  & \multicolumn{3}{c}{\textbf{Video Quality}}                  \\
& \textbf{RotErr$\downarrow$} & {\textbf{TransErr$\downarrow$}} & {\textbf{CLIP-score$\uparrow$}} & \textbf{Smoothness$\uparrow$} & \textbf{Dynamic$\uparrow$} & \textbf{Aesthetic$\uparrow$} \\ \hline
ReCamMaster-Wan~\cite{bai2025recammaster}& 1.454& 5.695& 0.219& 0.917& \textbf{31.751}& 4.738\\
Ours& \textbf{1.275}& \textbf{5.667}& \textbf{0.220}& \textbf{0.933}& 24.21& \textbf{4.826}\\\hline

\end{tabular}}
\end{table}

\section{Experiments}
\paragraph{Dataset and Benchmark}

UNIC is trained on multiple datasets to support multi-task video editing. For ID swap/insert/delete and stylization tasks, we use self-constructed datasets (see Appendix for details). These are also adapted for the propagation task by using the first frame of the target video as the edited input. For re-camera control, we use the Multi-Cam Video Dataset from ReCamMaster~\cite{bai2025recammaster}. To comprehensively evaluate the method, we construct \textbf{a unified video editing benchmark} incorporating six representative video editing tasks with distinct editing area ratio and conditioning modalities. The details of the benchmark can be found in Appendix. 

\paragraph{Evaluation Metrics}
The evaluation of these tasks is conducted across two primary dimensions: task-specific performance and overall video quality. To assess task-specific performance, for ID tasks, we employ the DINO-score~\cite{caron2021dino} and CLIP-score~\cite{clip} to evaluate identity similarity with the reference image. For the style task, style similarity with the reference style image is validated using the CSD-score~\cite{csd}, while ArtFID and CFSD~\cite{styleid} are used to consider content preservation. Furthermore, for re-camera control, we use RotErr, TransErr, and CamMC following CamI2V~\cite{zheng2024cami2v} to evaluate the alignment with the given camera pose. Besides, the overall quality of the generated videos is assessed by smoothness, dynamic quality, and aesthetic quality~\cite{huang2024vbench}.

\subsection{Comparisons}

We compare our approach with the state-of-the-art unified video editing methods like VACE~\cite{jiang2025vace} and task-specific methods from different video editing methods, like ReCamMaster~\cite{bai2025recammaster}. We provide an evaluation of overall video performance, including text alignment and overall quality, as well as task-specific metrics such as DINO-score in ID-related tasks. Since the propagation version (indicated by a (prop) suffix) for Swap/Delete/Stylization tasks requires the edited first frame, we specifically use Insert-Anything~\cite{song2025insert}, FLUX~\cite{flux}, and the first frame of StyleMaster~\cite{ye2024stylemaster} to obtain the edited first frame.

As presented in Table~\ref{maintable}, our six-in-one framework demonstrates consistent and strong performance across all evaluated video editing tasks. Notably, our model achieves leading results in ID Insert and Re-Camera Control, outperforming existing methods on most metrics. For Stylization, we achieve comparable performance to specialized models like StyleMaster~\cite{ye2024stylemaster}. While for ID Delete, specialized models like VideoPainter~\cite{bian2025videopainter} show better video reconstruction with PSNR, our approach still surpasses it in alignment with CLIP-score and some video quality aspects such as Smoothness and Aesthetic score. Furthermore, a significant advantage of our method is its flexibility, supporting arbitrary resolutions and lengths, a capability not present in many fixed-length video editing models.

\subsection{Analysis for Our In-Context Video Editing}
We provide a set of experiments to demonstrate the training choice of in-context video editing for different tasks. Based on our experimental results, we summarize several key findings that highlight how specific training strategies enable robust multi-task unification and enhance the overall performance of our in-context framework.

\paragraph{Should we train tasks sequentially or jointly?}
To determine the optimal training strategy for our unified tasks, which possess varying levels of difficulty, we investigate whether sequential training or joint training yields superior performance. As shown in Table~\ref{tab:training_order}, we report results under different training strategies. Among our selected six tasks, re-camera control is the most difficult, since the modality is far away from the visual content, while the ID-related task is relatively easy to learn. We experiment with three approaches: (1) sequential training from hard to easy tasks, (2) sequential training from easy to hard tasks, and (3) joint training of all tasks from scratch. Our findings indicate that the sequential training approaches (hard-to-easy and easy-to-hard) can help the multi-task learning. In contrast, joint training from scratch, while capable of learning easier tasks, struggles significantly with harder ones, resulting in poor performance on the re-camera control task.

\begin{table}[!t]
\centering
\caption{\textbf{Ablation study on the training order of different tasks.} We employ three settings: hard to easy, easy to hard, and joint training. Best results are highlighted in \textbf{bold}. $\uparrow$ indicates higher is better;  $\downarrow$ indicates lower is better.
}
\label{tab:training_order}
\resizebox{\textwidth}{!}{
\begin{tabular}{@{}l|cc|ccc|ccc@{}}
\toprule
\multirow{2}{*}{\textbf{Training Order}}  & \multicolumn{2}{c|}{\textbf{Identities}}& \multicolumn{3}{c|}{\textbf{ReCam}}& \multicolumn{3}{c}{\textbf{Style}}\\
 & \textbf{CLIP-I$\uparrow$}& \textbf{DINO-I$\uparrow$} & \textbf{RotErr$\downarrow$}  &\textbf{TransErr$\downarrow$}  & \textbf{CamMC$\downarrow$}& \textbf{CSD-Score$\uparrow$}  &\textbf{ArtFID$\downarrow$} & \textbf{CFSD$\downarrow$}\\
\midrule
\begin{tabular}[c]{@{}l@{}}{-> {camera}} \\ -> {camera+id} \\ -> {camera+id+style+propagation}\end{tabular} & 0.725& \textbf{0.429}&  \textbf{1.275}&\textbf{5.667}&  \textbf{6.154}&  0.259&\textbf{37.619}& \textbf{0.107}\\
\midrule
\begin{tabular}[c]{@{}l@{}}{-> {id}} \\ -> {id+style+propagation} \\ -> {camera+id+style+propagation}\end{tabular} & \textbf{0.726}& 0.427&  1.398&5.681&  6.275&  0.247&37.748& 0.109\\
\midrule
\begin{tabular}[c]{@{}l@{}}-> {camera+id+style+propagation}\end{tabular} & 0.713& 0.421&  2.287&9.694&  10.377& \textbf{0.298}&38.953& 0.170\\
\bottomrule
\end{tabular}}
\end{table}

\paragraph{Will the task unification affect single-task performance?}
There is a natural concern about whether multi-task learning affects single-task performance. To explore this, we conduct a comparative study between task-specific models and the unified model. We train separate models for stylization, ID-related editing, and re-camera control, and compare their performance with the unified model. As shown in Table~\ref{tab:singlevsunified}, the unified model does not impair task performance and even offers advantages in camera control and style similarity in stylization. However, it slightly compromises content preservation in stylization. This trade-off is due to the training mechanism: during ID-related tasks, the model is trained to fully preserve content. As a result, in stylization tasks, the unified model retains more information than just style, leading to higher style similarity but reduced content preservation. Overall, unifying diverse tasks within this framework does not significantly degrade individual task performance and can even enhance it in some cases.

\begin{table}[!t]
\centering
\caption{\textbf{Performance Comparison between task-specific model and unified model.} We evaluate our unified model (B4) against task-specific models trained for ID (B1), style (B2), and re-camera control (B3) on relevant metrics for each task. Best results are highlighted in \textbf{bold}. $\uparrow$ indicates higher is better;  $\downarrow$ indicates lower is better.
}
\resizebox{\textwidth}{!}{\begin{tabular}{@{}l|l|cc|ccc|ccc@{}}
\toprule
&\multirow{2}{*}{\textbf{Task}} & \multicolumn{2}{c|}{\textbf{Identities}}& \multicolumn{3}{c|}{\textbf{Style}}& \multicolumn{3}{c}{\textbf{ReCam}} \\
& & \textbf{DINO-I$\uparrow$}  &\textbf{CLIP-I$\uparrow$}& \textbf{CSD-Score$\uparrow$}  &\textbf{ArtFID$\downarrow$} & \textbf{CFSD$\downarrow$}& \textbf{RotErr$\downarrow$}  & \textbf{TransErr$\downarrow$} & \textbf{CamMC$\downarrow$}\\
\midrule
B1 & id & \textbf{0.449}&0.723&  -&-& -& -& -& -\\
B2 & style &  -&-& 0.234&37.674& \textbf{0.096}& -& -& -\\
B3 & camera&  -&-&  -&-& -& 1.472& 5.836& 6.434\\
B4 & id+style+camera& 0.429&\textbf{0.725}&  \textbf{0.259}&\textbf{37.619}& 0.107& \textbf{1.275}& \textbf{5.667}& \textbf{6.154}\\
\bottomrule
\end{tabular}}
\label{tab:singlevsunified}
\end{table}

\paragraph{Do condition bias and task-aware RoPE Matter?}
To validate our proposed Condition Bias and Task-Aware RoPE, we conduct an ablation study, with results presented in Table~\ref{tab:ablation_components}. Comparing our full model against variants clearly demonstrates their benefits. The baseline model, D1, which lacks both designs, can handle relatively easy tasks but struggles significantly with hard tasks requiring precise per-frame alignment, such as re-camera control, exhibiting high TransErr and CamMC. Instead, introducing Condition Bias alone in model D2 offers improvements, for instance, boosting CLIP-I in ID swap and reducing TransErr for re-camera control. However, it still shows limitations in other temporal aspects. Conversely, model D3, which employs Task-aware RoPE without Condition Bias, shows a marked improvement specifically in re-camera control, largely reducing CamMC. This highlights its strength in handling complex temporal alignments. The full model, D4, incorporating both designs, achieves the best overall performance. It significantly outperforms the baseline D1 across all task categories. This synergistic effect underscores that Condition Bias provides crucial task disambiguation, while Task-aware RoPE offers robust temporal modeling, both vital for a unified and effective video editing framework.

\begin{table}[!t]
\centering
\caption{\textbf{Ablation Study on Condition Bias and Task-aware RoPE.} We compare our full model (D4) with their variants. Best results are highlighted in \textbf{bold}. $\uparrow$ indicates higher is better;  $\downarrow$ indicates lower is better.}

\resizebox{\textwidth}{!}{\begin{tabular}{@{}l|c|cc|cc|ccc|ccc@{}}
\toprule
& \textbf{Condition} &\multicolumn{2}{c|}{\textbf{RoPE}}& \multicolumn{2}{c|}{\textbf{Identities}}& \multicolumn{3}{c|}{\textbf{Style}}&\multicolumn{3}{c}{\textbf{ReCam}}\\
&  \textbf{Bias}&  \textbf{Sequential}&\textbf{Task-aware}& \textbf{DINO-I$\uparrow$}  &\textbf{CLIP-I$\uparrow$}& \textbf{CSD-Score$\uparrow$}  &\textbf{ArtFID$\downarrow$} & \textbf{CFSD$\downarrow$}& \textbf{RotErr$\downarrow$}  & \textbf{TransErr$\downarrow$} & \textbf{CamMC$\downarrow$}\\
\midrule
D1& &  \checkmark& & 0.433& 0.710& 0.242& 34.194& 0.081& 2.501& 8.972& 13.119\\
D2& \checkmark&  \checkmark& & \textbf{0.434}& 0.723& \textbf{0.274}& 35.548& 0.091& 1.428& 6.039& 6.566\\
D3& &  & \checkmark& 0.422& 0.710& 0.258& \textbf{32.768}& \textbf{0.072}& 1.304& 6.038& 6.498\\
D4& \checkmark&  & \checkmark& 0.429& \textbf{0.725}& 0.259& 37.619& 0.107& \textbf{1.275}& \textbf{5.667}& \textbf{6.154}\\
\bottomrule
\end{tabular}}
\label{tab:ablation_components}
\end{table}

\section{Conclusion}
In this paper, we introduce UNified In-Context Video Editing (UNIC), a simple yet effective framework that unifies diverse video editing tasks within a
single model in an in-context manner. To this end, we formulate the input of different video editing tasks as three types of tokens, integrating them as a single unified token sequence jointly modeled with the original full-attention of diffusion transformers. With the devised task-aware RoPE and conditional bias, our method can flexibly perform different editing tasks and support their combination. To facilitate the evaluation, we also construct a unified video editing benchmark. Extensive experiments on six representative video editing tasks demonstrate that our unified model shows superior performance on each task and exhibits emergent task composition abilities.

\bibliographystyle{unsrt}
\bibliography{main}


\renewcommand{\figurename}{Fig.S\kern -2.5pt}
\title{Unified In-Context Video Editing}

%

\author{%
  David S.~Hippocampus\thanks{Use footnote for providing further information
    about author (webpage, alternative address)---\emph{not} for acknowledging
    funding agencies.} \\
  Department of Computer Science\\
  Cranberry-Lemon University\\
  Pittsburgh, PA 15213 \\
  \texttt{hippo@cs.cranberry-lemon.edu} \\
}

\newpage
\appendix
\section*{Appendix}
\addcontentsline{toc}{section}{Appendix}

\section{UNIC Benchmark}
To comprehensively evaluate performance, we create a unified benchmark of six tasks, each containing 20 to 50 carefully designed evaluation cases. 
\subsection{ID Insert}
\begin{figure}[h]
    \centering
    \includegraphics[width=0.6\linewidth]{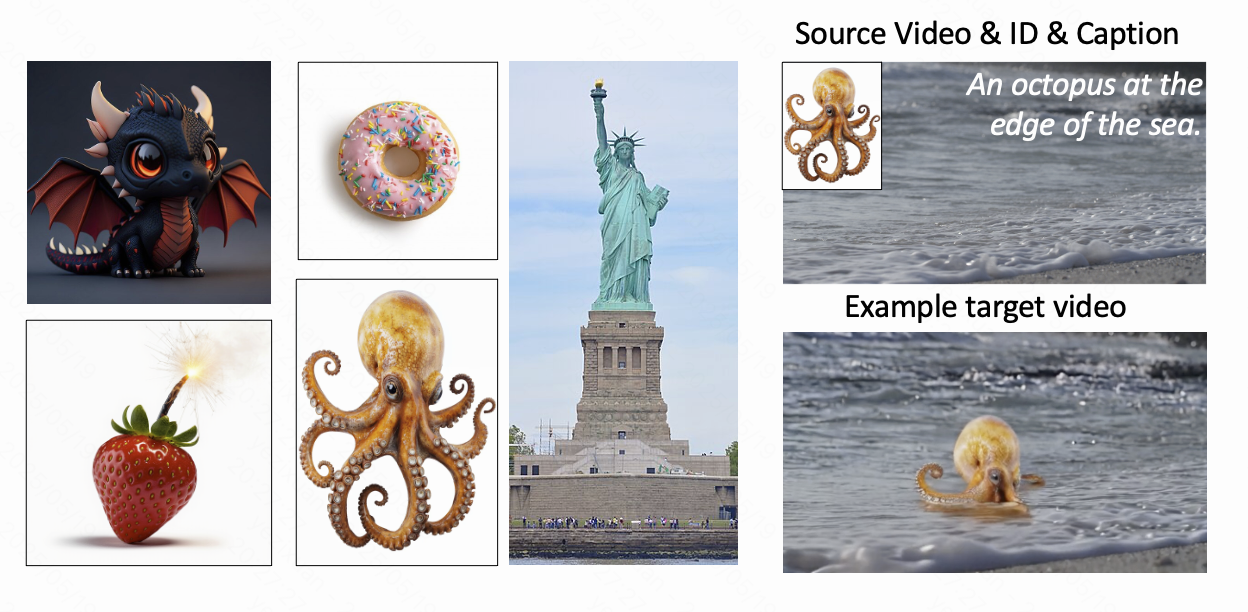}
    \caption{\textbf{ID Pool and example of ID insert evaluation cases.}}
    \label{fig:supp-id-example}
\end{figure}

We collect $20$ videos from Artgrid~\cite{artgrid} as source video, and we carefully select a suitable ID for each video to insert, also ensuring that the semantic is reasonable. As shown in Fig.S \ref{fig:supp-id-example}, our selected ID includes both clean object without background and complete picture with background.

\subsection{ID Swap}
\begin{figure}[h]
    \centering
    \includegraphics[width=1\linewidth]{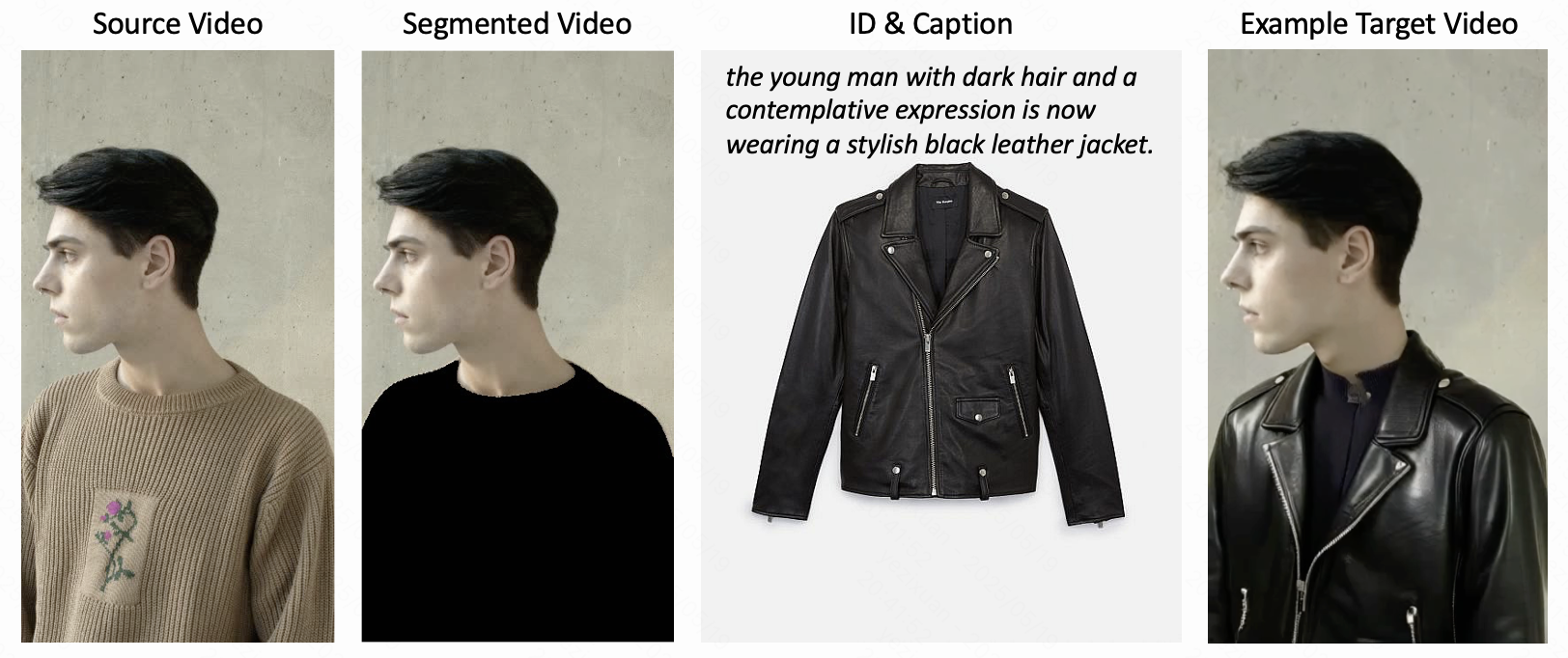}
    \caption{\textbf{Example of ID swap evaluation cases.}}
    \label{fig:supp-id-swap-example}
\end{figure}

For this task, we utilize $14$ videos from VPBench~\cite{bian2025videopainter} and $6$ videos online as the source videos. With the source videos, the objects to be swapped were segmented using SAM2~\cite{ravi2024sam2}. Then, we carefully design and choose the object to place in. As shown in Fig.S\ref{fig:supp-id-swap-example}, an appropriate ID was then selected from our ID pool to replace the segmented object, and a caption was generated to describe the final target video.

\subsection{ID Delete}
\begin{figure}[h]
    \centering
    \includegraphics[width=1\linewidth]{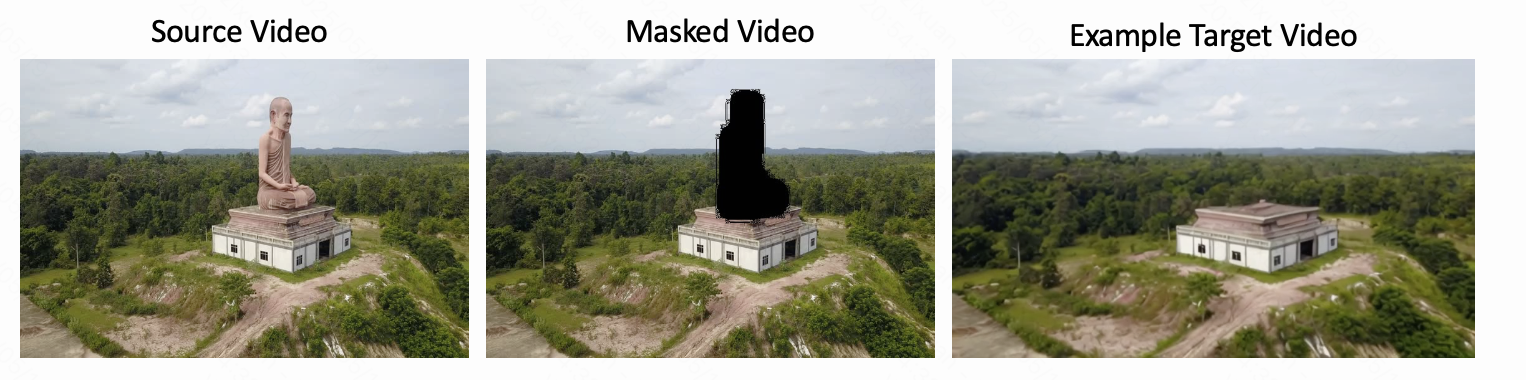}
    \caption{\textbf{Example of ID Delete evaluation cases.}}
    \label{fig:supp-id-delete-example}
\end{figure}

For ID delete, we expand the $10$ videos in VPBench~\cite{bian2025videopainter} to $20$ by additionally collecting $10$ videos. Also, SAM2~\cite{ravi2024sam2} is used to segment the object to be deleted. Then we generate caption for the target video. The example is shown in Fig.S\ref{fig:supp-id-delete-example}.

\subsection{Stylization}
\begin{figure}[h]
    \centering
    \includegraphics[width=1\linewidth]{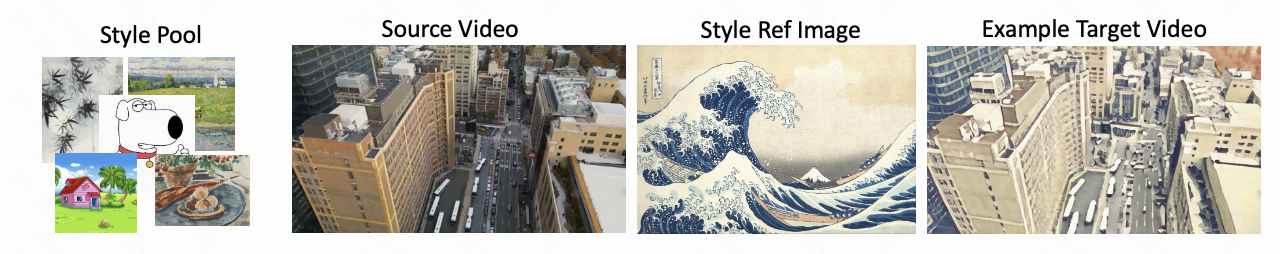}
    \caption{\textbf{Example of Stylization evaluation cases.}}
    \label{fig:supp-style-example}
\end{figure}

For the stylization task, we collected $12$ representative styles to serve as references. These include diverse artistic expressions such as pixel art, oil painting, Chinese painting, and line art, among others. Examples of these styles are illustrated in Fig.S\ref{fig:supp-style-example}. Then we randomly select $50$ videos from Artgrid~\cite{artgrid} as the source videos.

\subsection{Propagation}
\begin{figure}[h]
    \centering
    \includegraphics[width=1\linewidth]{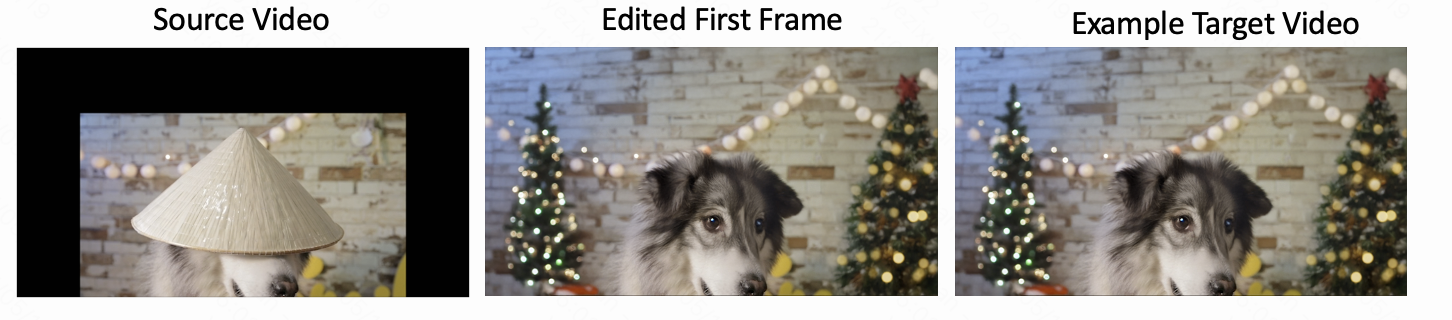}
    \caption{\textbf{Example of propagation evaluation cases.}}
    \label{fig:supp-prop-example}
\end{figure}

For the propagation task, we expand the $38$ example of GenProp~\cite{liu2024generative} to $50$ examples by adding $12$ stylization propagation test cases. Example is shown in Fig.S\ref{fig:supp-prop-example}.
\subsection{Re-camera Control}
To evaluate the re-camera control task, we utilized $10$ basic camera trajectories and $50$ randomly selected videos from Koala~\cite{wang2024koala}. Each of the $10$ trajectories was then applied to $5$ distinct videos from this set (totaling $50$ trajectory-video pairs).

\section{Training Dataset Construction}
This section details the construction of our datasets for the six tasks.
\subsection{ID-related Task}
\begin{figure}[h]
    \centering
    \includegraphics[width=1\linewidth]{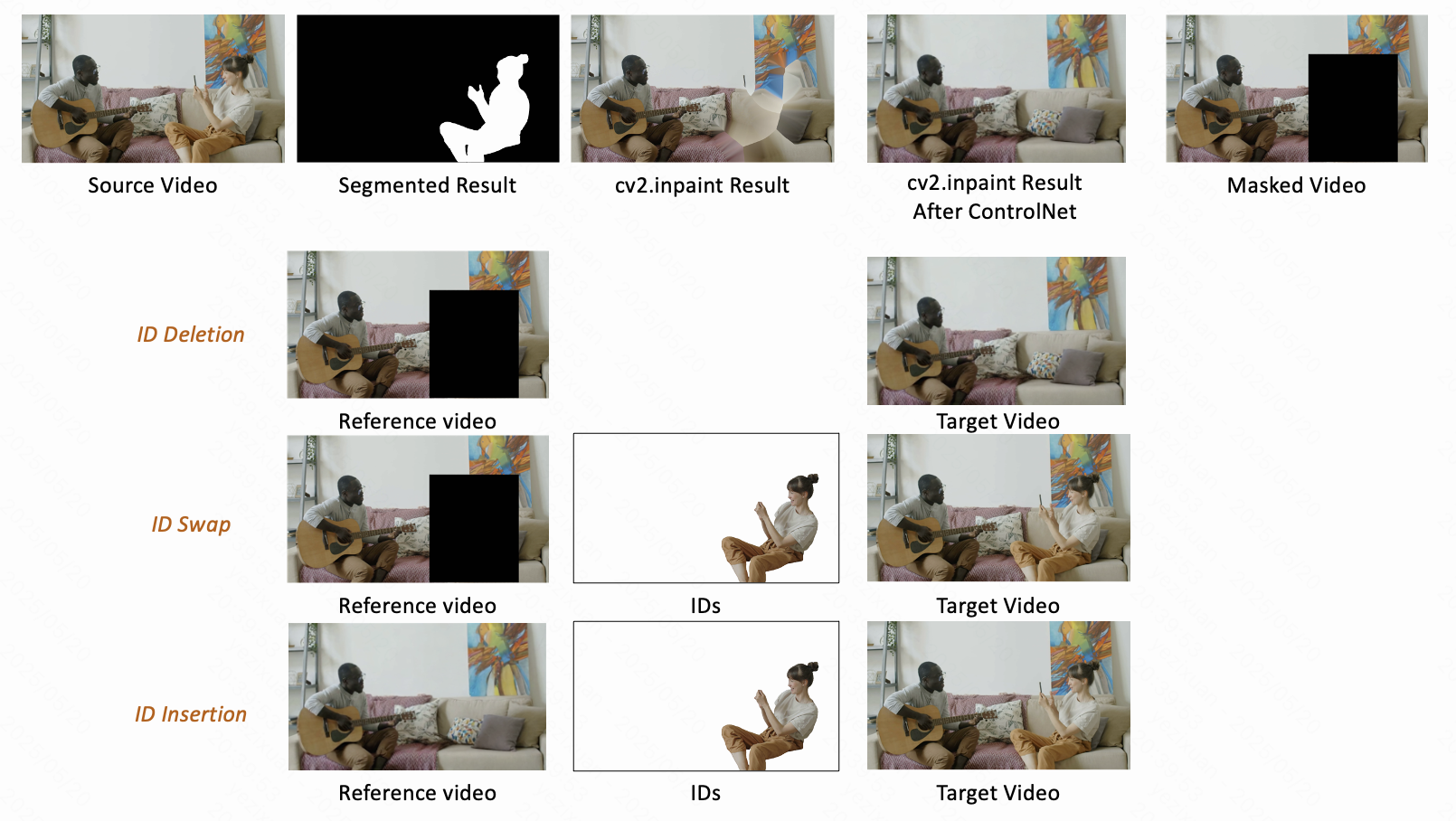}
    \caption{\textbf{ID-related task data construction.}}
    \label{fig:supp-id-data}
\end{figure}

To generate training data for ID-related tasks such as deletion, swap, and insertion (as illustrated in Fig.~S\ref{fig:supp-id-data}), we first use SAM2~\cite{ravi2024sam2} to obtain an object's segmentation mask from the source video.
This mask is then applied with \texttt{cv2.inpaint} to produce an inpainted video.
However, this simple inpainting method often introduces visual artifacts in the inpainted regions.
To address this, we train a ControlNet conditioned on the original video, which effectively eliminates these artifacts.
The resulting artifact-free video serves as the reference video for the insertion task, with the original source video as the target.
Additionally, a masked video, created by applying the segmentation mask to the source video, serves as the reference video for both deletion and swap tasks.

\subsection{Stylization}

When considering how to construct the paired style video dataset, a straightforward idea is to use a video-to-video stylization model to convert real-world videos into stylized ones. However, our experiments revealed that this approach frequently results in temporal inconsistencies, flickering artifacts, and lower visual quality.

\begin{figure}[h]
    \centering
    \includegraphics[width=\linewidth]{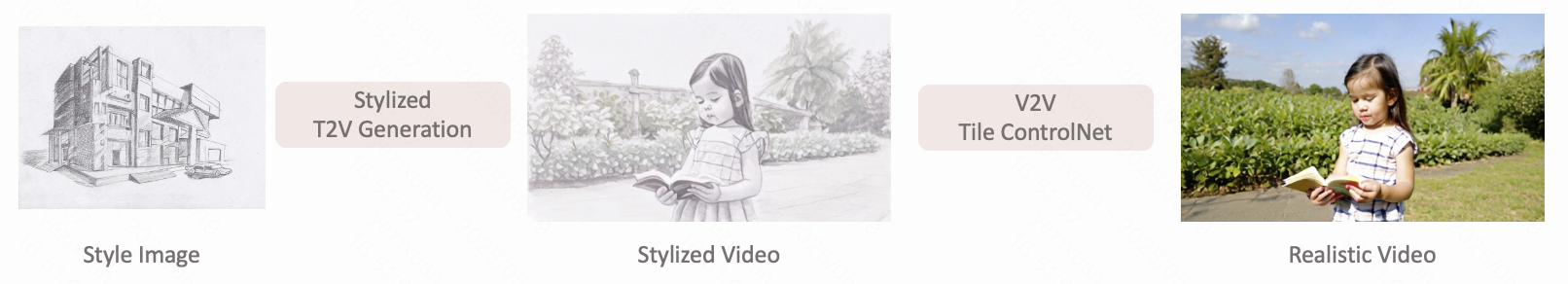}
    \caption{\textbf{Stylization pair data construction.}}
    \label{fig:supp-style-dataset}
\end{figure}

We then noted that Text-to-Video (T2V) models are capable of generating stylized videos that exhibit superior quality and maintain higher fidelity to a given reference style image. This observation led us to an alternative strategy: rather than stylizing an existing real video, we first generate a high-quality stylized video using a T2V model. Subsequently, we transform this stylized video into a realistic counterpart using a tile-based video ControlNet. As illustrated in Fig.S\ref{fig:supp-style-dataset}, the results confirm that this is a feasible method. Using this method, we create 10,000 paired videos.

\subsection{Propagation Task Dataset}
To construct a dataset for the propagation task, we leverage existing paired data from our ID-related and Stylization task training sets. Each pair in these datasets contains a source video and a target video.

The propagation task requires input triplets consisting of a source video, a target video, and the first frame of that target video. We can generate these triplets from our existing paired data in two distinct configurations. In one configuration, the original source video serves as the propagation source, the original target video serves as the propagation target, and we use the first frame of this original target video. Alternatively, the roles can be reversed: the original target video can serve as the propagation source, with the original source video becoming the propagation target, and we would then use the first frame of this (now) target video. This strategy effectively allows us to derive a greater volume of propagation training data from our existing resources.

\subsection{Re-Camera Control Task Dataset}
For the Re-Camera control task, we employ the Multi-Cam Video dataset from the ReCamMaster~\cite{bai2025recammaster}. This established training dataset provides 136,000 videos.

\section{Training Schemes}
\subsection{Model Details for Six Tasks}
\begin{figure}[h]
    \centering
    \includegraphics[width=1\linewidth]{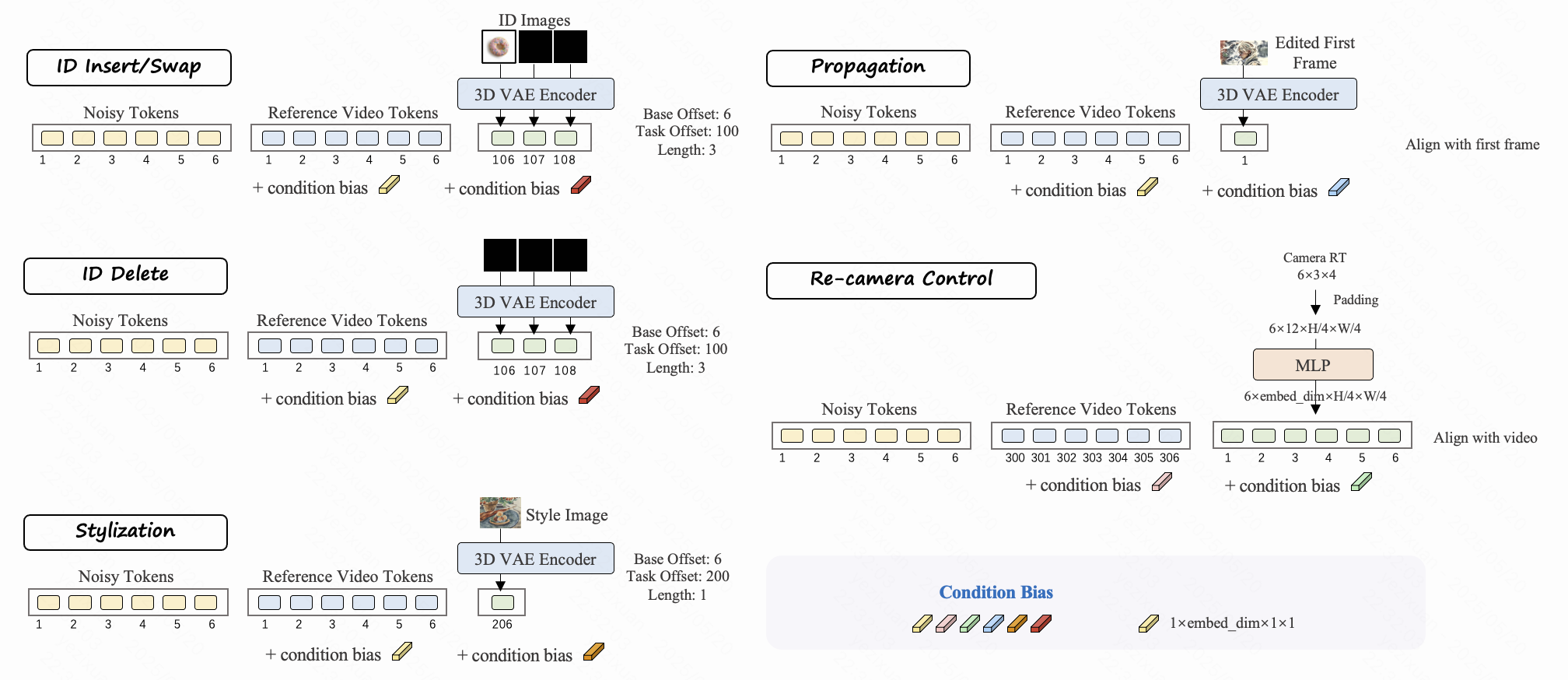}
    \caption{\textbf{Detailed Input and RoPE index for the six tasks.}}
    \label{fig:supp-arc-detail}
\end{figure}

Figure~S\ref{fig:supp-arc-detail} illustrates the model architecture details for handling the six distinct tasks. The configuration for each task, particularly concerning input encoding and Rotary Positional Encoding (RoPE) indices, is as follows:

\begin{itemize}
    \item \textbf{ID Insertion and Swapping:}
    For these tasks, the injected ID image(s) are encoded using the same 3D VAE Encoder employed for the reference video.
    Their RoPE indices commence from a base offset of $6$, with an additional task-specific offset of $100$. This allocation has a length of $3$, accommodating up to three ID images.
    If fewer than three ID images are provided for a given instance, the embedding slots for the remaining IDs are filled with representations corresponding to black images.

    \item \textbf{ID Deletion:}
    In the ID deletion task, all input ID image slots are effectively treated as black images (i.e., their embeddings correspond to black images), signaling the removal operation. The RoPE indexing follows the same base offset of $6$ and task offset of $100$ as ID insertion/swapping.

    \item \textbf{Stylization:}
    For the stylization task, the style reference image is also embedded using the 3D VAE Encoder.
    It utilizes the same base RoPE offset of $6$, but with a different task-specific offset of $200$.
    Therefore, the RoPE indices for style tokens begin at $206$ (i.e., $6 + 200$).

    \item \textbf{Propagation:}
    The propagation task leverages the direct correspondence with the first frame of the target video.
    Consequently, the tokens representing this first frame (used as the propagation source/reference) are assigned RoPE index $1$.

    \item \textbf{Re-camera Control:}
    Camera parameters for re-camera control, initially provided as a tensor of size $F \times 3 \times 4$ (where $F$ is the number of frames), undergo a specific tokenization process.
    First, the last two dimensions ($3 \times 4$) are flattened, resulting in $F \times 12$.
    These features are then spatially padded to match the VAE token dimensions of $H/4 \times W/4$ to obtain $F \times 12 \times H/4 \times W/4$.
    Subsequently, an MLP embeds these processed parameters into a tensor of shape $F \times \text{emb\_dim} \times H/4 \times W/4$, aligning them with the dimensionality of tokens encoded by the 3D VAE.
    Crucially, since this task regards the reference video primarily as soft guidance rather than requiring strict pixel-to-pixel alignment, the RoPE indices for the reference video tokens are shifted by $+300$ from their original positions (e.g., original index $i$ becomes $i+300$).
\end{itemize}

\subsection{Training Progress}
\begin{figure}[h]
    \centering
    \includegraphics[width=1\linewidth]{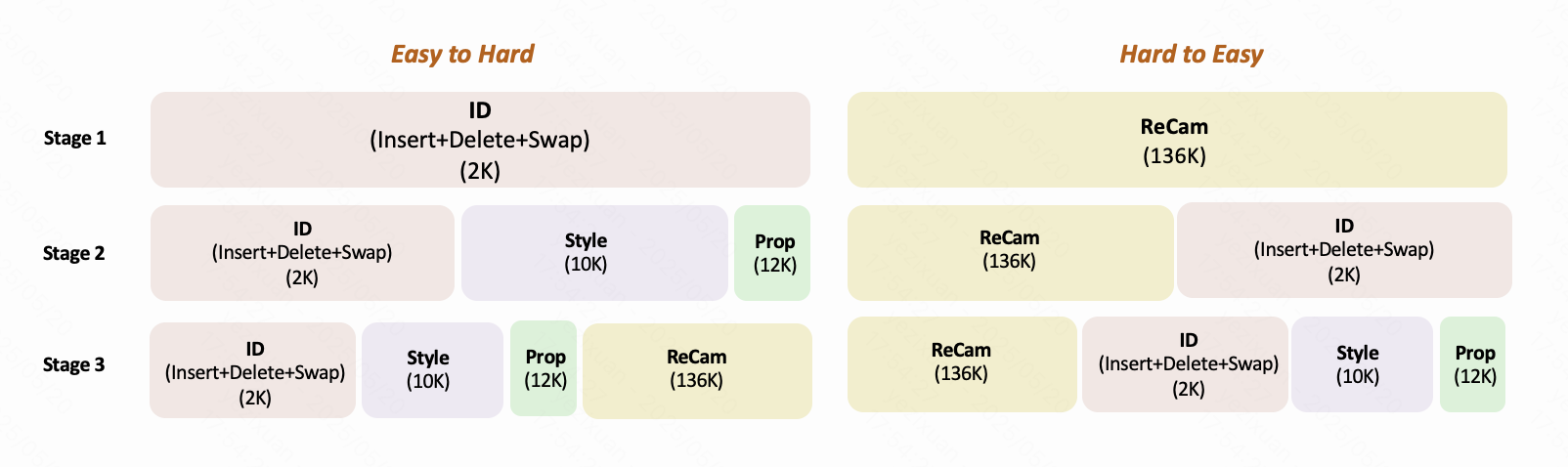}
    \caption{\textbf{Training progress with different settings.} We demonstrate two settings of the progress and the data volume of each task.}
    \label{fig:supp-training-progress}
\end{figure}

As shown in Fig.S\ref{fig:supp-training-progress}, we present two training progression strategies for our method: one starting with tasks deemed ``hard'' and progressing to ``easy'' ones, and the converse strategy, from ``easy'' to ``hard''.
The classification of tasks as ``hard'' or ``easy'' is based on our empirical observations during training within this in-context framework. For instance, we found that the re-camera control task typically requires training on approximately $100\text{k}$ data volume to reach satisfactory performance, whereas ID-related tasks (such as ID insertion or swapping) achieve comparable results with only about $20\text{k}$ volume.
Consequently, we categorize re-camera control as a ``harder'' task in this context.

\subsection{Training Details}
UNIC is trained on multiple datasets to support multi-task video editing. All finetuning experiments start from a pre-trained model with 1B parameters and 28 sequential standard Diffusion Transformer (DiT) blocks. The model is finetuned for 16k iterations on 32 H800 GPUs with a batch size of 64. We only finetune the transformer module and the new tokenizers (like the MLP for the camera pose), while freezing the 3D VAE, Text T5 Tokenizer.

\section{Limitation and Future Work}
Our current unification efforts are limited to the six tasks discussed. The incorporation of additional modalities, such as lip-syncing with audio, remains unexplored in this work. In future research, we intend to integrate a wider array of tasks to investigate whether there is an upper limit to the number of tasks that can be successfully unified. Furthermore, recognizing that a high token count during self-attention can significantly increase computational overhead, we plan to explore architectural designs or alternative mechanisms to improve efficiency.

\end{document}